\documentclass[10pt,twocolumn,letterpaper]{article}

\usepackage[pagenumbers]{cvpr} 

\usepackage[ruled,vlined]{algorithm2e}

\usepackage{amsmath, amssymb, mathtools}

\usepackage{dsfont}   

\usepackage{tabularx} 
\usepackage{booktabs} 
\usepackage{multirow} 
\usepackage{array}    
\usepackage{makecell}

\usepackage{graphicx}
\usepackage{xcolor}
\usepackage{siunitx}
\usepackage{caption}

\definecolor{cvprblue}{rgb}{0.21,0.49,0.74}
\usepackage[pagebackref,breaklinks,colorlinks,allcolors=cvprblue]{hyperref}

\usepackage{soul}

\def\paperID{6753} 
\def\confName{CVPR}
\def\confYear{2026}

\title{Towards Robust Protective Perturbation against DeepFake Face Swapping}

\author{
Hengyang Yao\thanks{Equal contribution.} \\
University of Birmingham\\
Birmingham, UK\\
{\tt\small h.yao.1@bham.ac.uk}
\and
Lin Li\footnotemark[1] \\
University of Oxford\\
Oxford, UK\\
{\tt\small lin.li@cs.ox.ac.uk}
\and
Ke Sun\\
Xiamen University\\
Xiamen, China\\
{\tt\small skjack@stu.xmu.edu.cn}
\and
Jianing Qiu\\
MBZUAI\\
Abu Dhabi, UAE\\
{\tt\small Jianing.Qiu@mbzuai.ac.ae}
\and
Huiping Chen\thanks{Corresponding author.}\\
University of Birmingham\\
Birmingham, UK\\
{\tt\small h.chen.13@bham.ac.uk}
}

\begin{document}
\maketitle

\begin{abstract}
DeepFake face swapping enables highly realistic identity forgeries, posing serious privacy and security risks. A common defence embeds invisible perturbations into images, but these are fragile and often destroyed by basic transformations such as compression or resizing. In this paper, we first conduct a systematic analysis of 30 transformations across six categories and show that protection robustness is highly sensitive to the choice of training transformations, making the standard Expectation over Transformation (EOT) with uniform sampling fundamentally suboptimal. Motivated by this, we propose Expectation Over Learned distribution of Transformation (EOLT), the framework to treat transformation distribution as a learnable component rather than a fixed design choice. Specifically, EOLT employs a policy network that learns to automatically prioritize critical transformations and adaptively generate instance-specific perturbations via reinforcement learning, enabling explicit modeling of defensive bottlenecks while maintaining broad transferability.  Extensive experiments demonstrate that our method achieves substantial improvements over state-of-the-art approaches, with 26\% higher average robustness and up to 30\% gains on challenging transformation categories.
\end{abstract}
\section{Introduction}
\label{sec:intro}

\begin{figure}
    \centering
    \includegraphics[width=\linewidth]{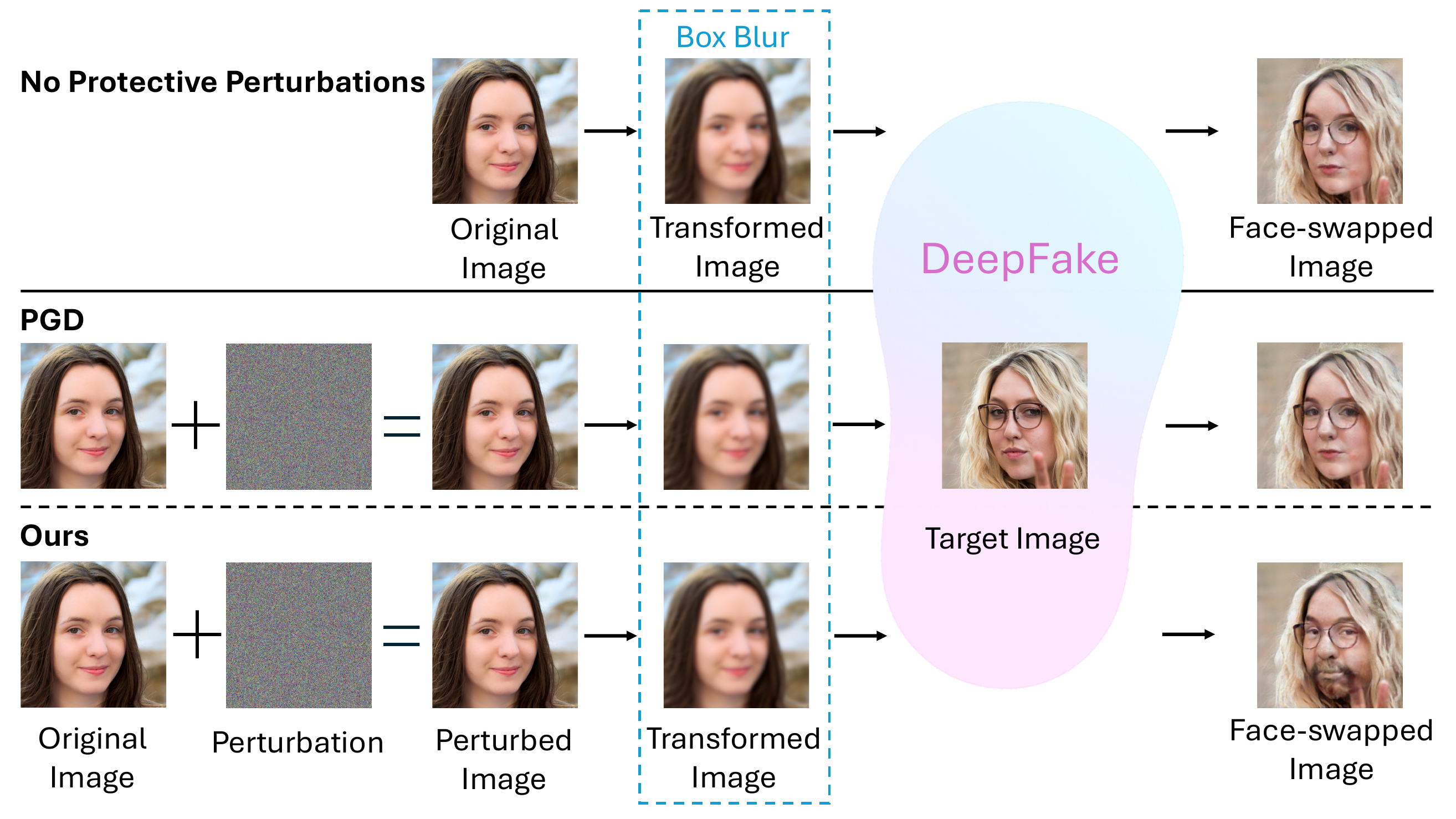}
    \caption{\textbf{Illustration of the research problem and our method.} The protective perturbation generated by the naive PGD algorithm \cite{madry_towards_2018} loses its effectiveness after the perturbed image undergoes box blur transformation, whereas our method maintains its protective effect and continues to disrupt the face-swapping output. Note that naive PGD achieves a similar protective effect as ours when no transformation is applied.}
    \label{fig: high level overview}
\end{figure}

DeepFake~\cite{gonzalez2018facial,huang2022fakelocator,nirkin2019fsgan,thies2016face2face,korshunov2018deepfakes}-synthetic media created using advanced generative models-have rapidly advanced in both realism and accessibility, enabling seamless manipulation of visual and auditory content. Among various forms of DeepFake, face swapping~\cite{chen2020simswap, simswap_plus, kim2025diffface} is one of the most widespread and impactful, allowing a person’s facial identity to be replaced with another’s while maintaining expressions, pose, and scene context. While this technology has enabled creative applications in entertainment and data augmentation, it also raises serious ethical and security concerns, such as identity theft, misinformation, and the erosion of digital trust.

To mitigate these threats, researchers have explored two main defense paradigms: passive and proactive defenses. Passive defenses \cite{sun2021dual,qian2020thinking,sun2022information,luo2023beyond,luo2023beyond,chen2022self,shiohara2022detecting} aim to detect forged content after creation by identifying spatial, temporal, or frequency inconsistencies, but they often struggle with generalization and are vulnerable to unseen manipulation methods. In contrast, proactive defenses \cite{huang2021initiative, aneja2022tafim, wang2022deepfake, salman2023raising} act preemptively by embedding protective perturbations into original images to prevent malicious manipulation. This approach is related to the longstanding challenge of adversarial examples in deep neural networks \cite{goodfellow_explaining_2015,madry_towards_2018,li_data_2023}, though with a different attack objective. A key limitation of passive defenses is that, in most cases, the harm has already occurred by the time manipulated images are detected. In comparison, protective perturbations secure the content at its source, helping to prevent negative consequences from arising.

However, existing proactive defenses face a severe limitation: their effectiveness often degrades under input transformations \cite{fu2022robust,qu_df-rap_2024,liu2024metacloak} such as compression, resizing, and color adjustment, which naturally occur during real-world data sharing or may be intentionally applied by malicious actors as countermeasures. Despite its importance for real-world deployment, this issue has been largely underexplored, and the community’s understanding remains limited. So far, the most effective approach to improving the robustness of protective perturbations is Expectation Over Transformation (EOT) \cite{athalye_synthesizing_2018}, which integrates a uniform distribution of transformations and aggregates perturbations generated from randomly sampled transformations.

In this work, we take a step toward addressing this limitation by conducting the first comprehensive investigation into the generalization of protective perturbations across 30 transformations. Our analysis reveals two critical findings: \textbf{(1) Transformations vary widely in generalization.} Some produce perturbations that transfer robustly across different categories of transformation, while others lead to severe overfitting or even reduce overall effectiveness. \textbf{(2) Certain transformations act as defensive bottlenecks.} Some cannot be adequately defended against through transfer from other transformation types and must be explicitly included during perturbation generation. These findings expose a fundamental flaw in the uniform random sampling approach used in EOT: it treats all transformations equally, overlooking their highly variable impact on robustness and failing to prioritize critical defensive bottlenecks.

Motivated by the above insights, we propose a new approach, named \textbf{E}xpectation \textbf{O}ver \textbf{L}earned distribution of \textbf{T}ransformation (EOLT), to enhance the robustness of protective perturbations. EOLT replaces the uniform prior over transformations in EOT with an optimized transformation distribution. This distribution is modeled by a deep neural network, referred to as the policy model, and is optimized by learning the policy model parameters to maximize the robustness of protective perturbations generated using sampled transformations against a target set of transformations.
Extensive experiments demonstrate that our method achieves substantial improvements over state-of-the-art approaches, with 26\% higher average robustness and up to 30\% gains on challenging transformation categories.

Overall, our contributions are twofold:
\begin{itemize}
    \item We present a comprehensive analysis of proactive defense robustness under 30 common input transformations, systematically examining their cross-transformation generalization behaviors. This analysis reveals several practical insights into how different transformations interact and impact protection effectiveness.
    
    \item We propose a novel method for generating robust protective perturbations against DeepFake face swapping. Our approach achieves state-of-the-art robustness across a wide range of input transformations, supporting the practical deployment of proactive defense strategies in real-world scenarios.
\end{itemize}



\section{Related Works}

\textbf{Face Swapping}. Face swapping transfers the identity of a source face onto a target face while preserving the target's pose, expression, and lighting. Modern methods fall into two categories: GAN-based and diffusion-based approaches. GAN-based methods evolved from subject-agnostic models like FSGAN~\cite{nirkin2019fsgan} to multi-stage architectures such as FaceShifter~\cite{li2019faceshifter}, with SimSwap~\cite{chen2020simswap, simswap_plus} achieving a strong balance between identity preservation and attribute fidelity in a lightweight framework. Later works integrated StyleGAN priors~\cite{zhu2021one, xu2022styleswap} and 3D geometric constraints~\cite{wang2021hififace, li20233d} for enhanced robustness. Diffusion-based approaches, pioneered by DiffFace~\cite{kim2025diffface}, offer superior stability and quality. Subsequent works formulated face swapping as conditional inpainting~\cite{zhao2023diffswap} or self-supervised learning, with ReFace~\cite{baliah2025realistic} masking target faces during inference for swapped results. In this work, we evaluate our proactive defense against SimSwap and ReFace as representative GAN-based and diffusion-based systems to assess robustness across generative paradigms under input transformations.

\textbf{Adversarial Examples as Proactive Defense}.
In proactive DeepFake disruption, one crafts a perturbation that, when added to a benign face, causes any subsequent DeepFake generator to output a clearly corrupted image \cite{huang2021initiative, aneja2022tafim, wang2022deepfake, salman2023raising, jeong2024faceshield, wang2024simac, wang2025nullswap}.
Recent work thus focuses on robustifying adversarial perturbations against such transformations. \citet{fu2022robust} combine EOT with Robust Error-Minimising Noise, minimising expected loss over augmentations and estimating gradients by sampling-and-averaging transforms, keeping the protective noise effective under common transforms. 
\citet{qu_df-rap_2024} propose DF-RAP, which explicitly models social‑media compression during training. They train a GAN (ComGAN) to mimic the effect of various online-compression pipelines and incorporate it into the adversarial-generation loop. \citet{liu2024metacloak} introduce MetaCloak, a meta-learning framework that samples a pool of image transformations (e.g. blurring, JPEG, super-resolution) during training so that the resulting perturbation is transferable and robust to such transformations.

\textbf{Robust Adversarial Examples}.
Prior works in adversarial machine learning have explored generating examples robust to input transformations. \citet{athalye_synthesizing_2018} proposed the Expectation over Transformation framework, which optimizes perturbations over transformation distributions to produce examples invariant to geometric and photometric changes. \citet{eykholt_robust_2018} extended this idea to the physical world with RP2, crafting traffic-sign perturbations that remain effective under varying viewpoints, distances, and lighting. \citet{zheng_robust_2023} introduced PadvFace, modeling sticker, face, and environmental variations, and using a curriculum adversarial attack (CAA) to create stickers capable of both dodging and impersonation in real-world settings. \citet{sitawarin_demystifying_2022} further showed that many transformation-based defenses fail against adaptive, transformation-aware attacks, emphasizing the need for rigorous evaluation.
While these methods target image classifiers, their assumptions and objectives differ fundamentally from those of face-swapping systems. As such, adapting robust adversarial examples to generative models requires specialized approaches.

\section{Problem Formulation and Background}


This section outlines the formal setup of the problem and gives necessary background. 
Consider a face-swapping model $\mathcal{F}(\cdot,\cdot)$ that transfers the identity of a source face $x$ onto a target face $x_t$, resulting in a synthesized image $y = \mathcal{F}(x,x_t)$. 
Since no operation is applied to the target face during this process, the equation can be simplified to $y = \mathcal{F}(x)$.

Protective perturbation is designed to degrade the output quality of face-swapping models while remaining visually imperceptible when applied to the original image. The optimization objective can be expressed as:
\begin{equation}
\label{equ: perturbation optimization}
\arg\max_{\|\delta\|_p \leq \epsilon}
\mathcal{L}\!\left(
 \mathcal{F}(x + \delta), 
\mathcal{F}(x)
\right),
\end{equation}
where $\mathcal{L}(\cdot,\cdot)$ measures the distance between two images, and $\|\delta\|_p \leq \epsilon$ constrains the perturbation magnitude to ensure imperceptibility.
In essence, this formulation searches for a perturbation to $x$ such that the synthesized, face-swapped, image generated from the perturbed $x$ is as dissimilar as possible from the one generated from the unperturbed $x$.
Throughout this work, we adopt the common white-box threat model \cite{madry_towards_2018}, assuming full access to the target face-swapping model to allow computation of gradients for the optimization in \cref{equ: perturbation optimization}.

We assume a set of transformations, $S_p$, is available for perturbation generation, and another set, $S_t$, is used for evaluation. These two sets may be identical or partially overlapping, reflecting scenarios in which a benign actor has full or partial knowledge of a malicious actor's anti-protection methods, respectively. The objective is to generate a perturbation based on $S_p$ that maintains high performance under the transformations in $S_t$ as measured below:
\begin{equation}
    \mathbb{E}_{t\in S_t}\left[\mathcal{U}\left(\mathcal{F}\left(t\left(x + \delta\right)\right)\right)\right]
\end{equation}
where $\mathcal{U}$ is a utility function or performance metric measures the quality of face swapping.


The baseline method EOT addresses this by optimizing the expected adversarial objective over a uniform distribution of $S_p$, formulated as:
\begin{equation}
\label{equ: EOT perturbation optimization}
\arg\max_{\|\delta\|_p \leq \epsilon}
\mathbb{E}_{t \sim p(S_p)}\left[\mathcal{L}\!\left(
\mathcal{F}\left(t\left(x + \delta\right)\right), 
\mathcal{F}(x)
\right)\right].
\end{equation}
In practice, a finite-typically small-number of transformations with strengths are randomly sampled from this uniform distribution to approximate the objective. For more technical details on EOT, we kindly refer readers to its original paper \cite{athalye_synthesizing_2018}.

\section{Generalization of Protective Perturbation Across Transformations}


\label{sec:sensitivity_simswap}

Prior work has demonstrated that incorporating transformations during adversarial perturbation generation can improve the robustness of protective perturbation against real-world distortions~\cite{athalye_synthesizing_2018}. However, existing methods typically adopt uniform random sampling over a transformation pool, implicitly assuming that all transformations contribute equally to robustness. This raises a key question: \emph{How does the choice of transformations during adversarial perturbation generation affect the cross-transformation generalization of protective perturbation?} To answer this, we perform a comprehensive analysis, measuring how perturbations generated with each individual transformation generalize to a wide range of input transformations.

\begin{figure}[]
    \centering
    \includegraphics[width=\linewidth]{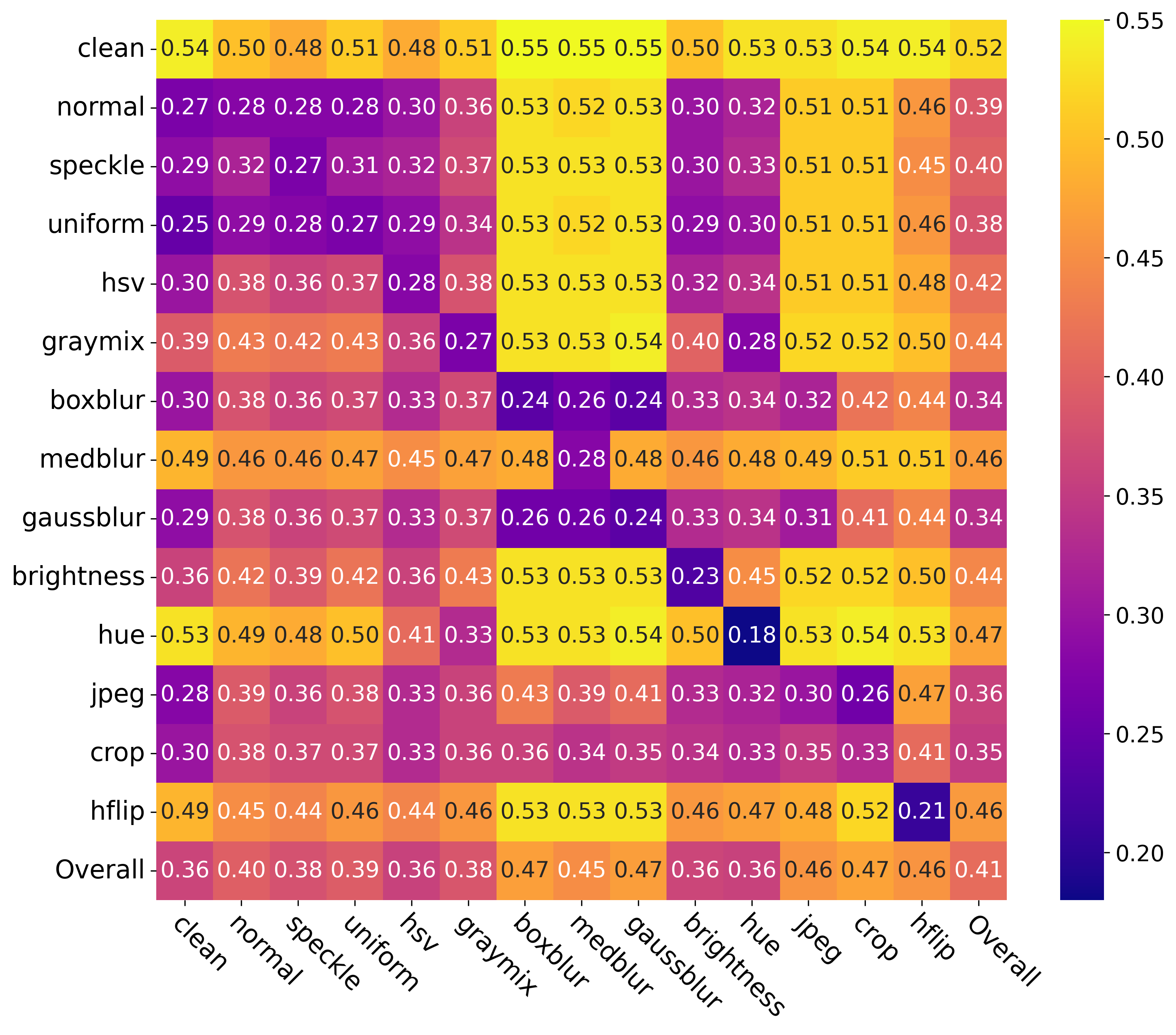}
    \caption{\textbf{Cross-transformation identity similarity matrix.} Rows indicate transformations used during perturbation generation, while columns indicate transformations applied at test time. ``Clean'' denotes no transformation applied.}
    \label{fig:simswap_heatmap}
\end{figure}

\textbf{Experimental setup.} We evaluate 30 transformations across six categories:
\textbf{Noise injection} (normal, uniform, speckle, poisson, salt, pepper),
\textbf{Color-space alteration} (hsv, lab, xyz, yuv, graymix),
\textbf{Blur filtering} (boxblur, medblur, motionblur, gaussblur),
\textbf{Stylization} (brightness, contrast, saturation, hue, gamma, solarize, sharp),
\textbf{Compression} (jpeg, fft, precision), and
\textbf{Geometric transforms} (affine, crop, hflip, vflip, swirl). Each transformation is associated with 9 intensity levels. Transformations that do not exhibit meaningful intensity variations (e.g., \textit{hflip}, \textit{vflip}) are duplicated across all 9 intensity slots to ensure a consistent number of logits per transformation, which is necessary for stable policy optimization. For each transformation $t$, we generate adversarial perturbations using PGD-EOT (100 steps, $\epsilon=0.05$) trained exclusively on $t$, then test against all 30 transformations plus the no-transformation baseline (``clean''). We measure identity similarity (lower is better) on 1,000 FFHQ pairs using SimSwap as the face-swapping model. 

Fig.~\ref{fig:simswap_heatmap} visualizes a representative subset of 15 transformations that illustrate key generalization patterns, with rows indicating training transformations and columns indicating test transformations (the complete 32$\times$32 matrix is provided in Appendix A). These results reveal two critical phenomena about how training and test-time transformations interact:

\textbf{Finding 1: Training transformations vary widely in cross-transformation generalization.}
We observe from Fig.~\ref{fig:simswap_heatmap} that some transformations induce broadly transferable perturbations, while others remain weak and highly specialized. Noise-type transformations (\texttt{normal}, \texttt{speckle}, \texttt{uniform}) demonstrate strong within-family generalization, maintaining ID similarities of 0.25--0.32 when tested against each other, indicating mutual transferability. Similarly, \texttt{boxblur} trained perturbations transfer well across categories, yielding consistently low ID similarities (0.24--0.38) on blur, noise, and color-space tests. By contrast, \texttt{hflip} trained perturbations perform well only on-diagonal but deteriorate off-diagonal to 0.44--0.53. Furthermore, \texttt{hue} trained perturbations average 0.47 across all tests, only marginally better than the unprotected baseline (0.52).
This heterogeneity indicates that transformation families exhibit vastly different transferability: some encode generalizable vulnerability patterns, while others induce severe overfitting.

\medskip
\textbf{Finding 2: Test-time transformations differ widely in defensive transferability.}
Some transformations are easily defended by perturbations trained on other distortions, while others remain hard targets unless explicitly seen during training. 
Column-wise patterns in Fig.~\ref{fig:simswap_heatmap} show that the blur family (\texttt{boxblur}, \texttt{medblur}, \texttt{gaussblur}) exhibits consistently high ID similarities (0.52--0.55) when perturbations are trained on non-blur transformations. For instance, noise-trained perturbations achieve 0.27--0.32 against noise tests but degrade to 0.52--0.53 against blur tests---an 80--90\% increase indicating near-complete defense failure. Only when blur itself appears during training do these columns darken to 0.24--0.28. Similarly, \texttt{crop} and \texttt{hflip} show elevated overall vulnerabilities (column averages of 0.47 and 0.46 respectively), indicating they are defensive bottlenecks that cannot be adequately covered via cross-transformation transfer and must be explicitly included in the training distribution.

\textbf{Implications.} To sum up, these findings challenge the common assumption that all transformations are equally beneficial for robustness and can be safely sampled uniformly in EOT. This suggests that the transformation distribution itself should be treated as a learnable component rather than a fixed design choice.

\section{Method}

Motivated by the above insights, we propose a new method, named \textbf{E}xpectation \textbf{O}ver \textbf{L}earned distribution of \textbf{T}ransformation (EOLT), which replaces the uniform prior with an optimized transformation distribution within the framework of EOT. The core idea is to parameterize the transformation distribution using a policy model and optimize its parameters to maximize the robustness of perturbations generated with the learned distribution against a target set of transformations. The policy model takes an image as input and outputs a probability distribution over a set of predefined transformations, allowing the transformation policy to be tailored to each individual source image.

\subsection{Parameterizing Transformation Policy}

\textbf{Transformation policy}.
We assume $K$ transformations are used for generating protective perturbations, \ie $S_p$. For most transformations, such as rotation and brightness, the range of valid strengths is either very large or continuous. Following prior works \cite{li_aroid_2024,cubuk_autoaugment_2019} in automated data augmentation, we discretize the strength of each transformation into $M$ magnitudes, allowing them to be modeled by a neural network as described below. 
Note that each transformation could have different number of magnitudes.
We refer to a specific transformation combined with a magnitude as a sub-policy. 
A transformation policy is then defined as a distribution over these sub-policies.

\begin{figure}
    \centering
    \includegraphics[width=\linewidth]{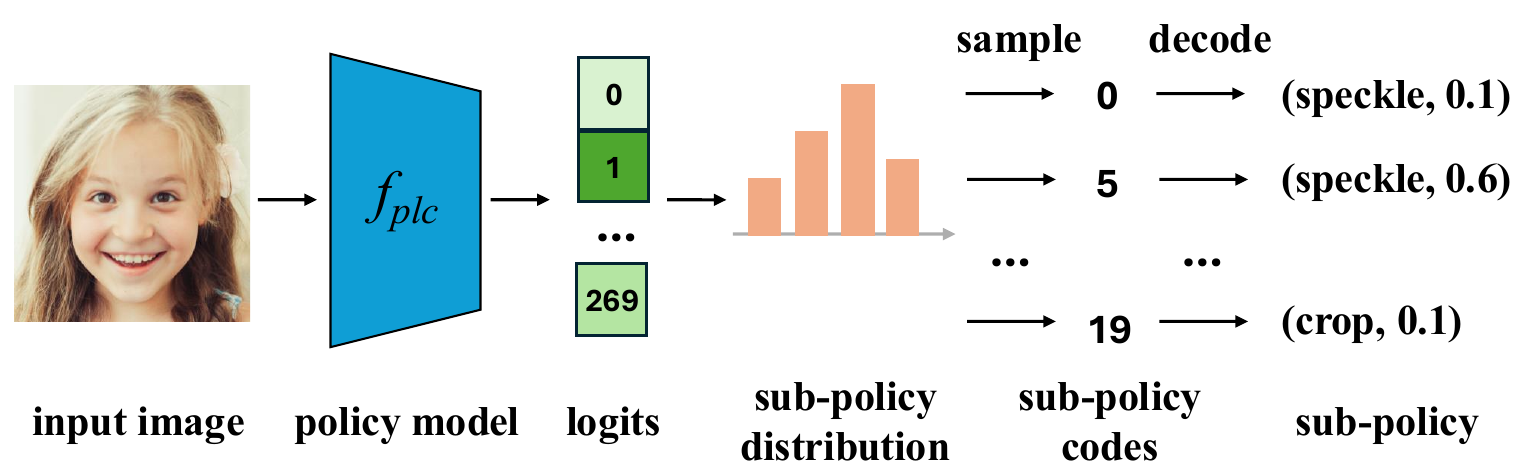}
    \caption{\textbf{Illustration of sampling multiple sub-policies from the policy model for a given image.} Taking an input image, the policy model produces a logits vector. These logits are converted into a sub-policy probability distribution, from which multiple sub-policy indices are sampled. Each sampled index is then decoded into a transformation-intensity combination, forming the sub-policies used during EOLT optimization.}
    \label{fig: policy sampling}
\end{figure}

\textbf{Policy modeling.} To optimize transformation policy efficiently, we model it as the output of a policy model parameterized by $\theta$ as illustrated in \cref{fig: policy sampling}. This model uses a deep neural network backbone to extract features from the input data, followed by a linear prediction head that generates the policy. The output of the head is interpreted as a multinomial distribution, where each logit corresponds to a predefined sub-policy. Different magnitudes of the same transformation are represented by separate logits, allowing each to be sampled independently. By conditioning the policy on the input data instance, the model can generate instance-specific policies tailored to the unique transformation needs of each input.

\subsection{Perturbation Generation}
Given a source image $x$ to protect, our policy model produces a transformation policy $\pi_{S_p}(x; \theta)$ over the transformation set $S_p$. We then replace the uniform distribution in Eq.~\ref{equ: EOT perturbation optimization} with this learned policy:
\begin{equation}
\label{equ: EOLT perturbation optimization}
\arg\max_{\|\delta\|_p \leq \epsilon}
\mathbb{E}_{t \sim \pi_{S_p}(x; \theta)}\left[\mathcal{L}\!\left(
\mathcal{F}\left(t\left(x + \delta\right)\right), 
\mathcal{F}(x)
\right)\right].
\end{equation}

In practice, we adopt the Projected Gradient Descent (PGD) algorithm \cite{madry_towards_2018} to solve the above optimization problem. 
PGD iteratively updates the perturbation $\delta$ with a fixed step size, $\alpha$, in the direction of the sign of the gradient of the adversarial objective, $\mathcal{L}_{\mathrm{adv}}$, from \cref{equ: EOLT perturbation optimization} with respect to the input image $x$, projecting $\delta$ back into an $\epsilon$-ball in $\ell_p$-norm to ensure visual imperceptibility. The iterative update rule is given by
\begin{equation}
\label{eq:pgd_update}
\delta^{(n+1)} = 
\text{Proj}_{\|\delta\|_p \le \epsilon}
\!\Big(
\delta^{(n)} + \alpha\, 
\text{sign}\!\big(\nabla_{x}\mathcal{L}_{\mathrm{adv}}\big)
\Big),
\end{equation}
\begin{equation}
\label{equ:pgd_xadv_short}
x_{adv}^{(n+1)} = x_{adv}^{(n)} + \delta^{(n+1)}, 
\qquad x_{adv}^{(0)} = x.
\end{equation}
where $n$ is the iteration index, $\text{sign}(\cdot)$ is a function returns the sign of the input, and $\text{Proj}_{\|\delta\|_\infty \le \epsilon}$ denotes projection onto the $\ell_\infty$ ball of radius $\epsilon$ centered at $x$.
Note that transformations are sampled from the learned policy in each iteration, ensuring that the sampled transformations vary across iterations.
After $N$ iterations, we obtain the final adversarial source image $x^{(N)}_{adv}$.

\subsection{Optimizing Transformation Policy}
The transformation policy is optimized by updating the policy model's parameters to maximize a target objective. This objective ensures that the perturbation generated based on the learned transformation policy (as defined in \ref{equ: EOLT perturbation optimization}) remains effective in disrupting face-swapping quality under a set of target transformations, denoted as $S_v$, analogous to a validation set in conventional machine learning. This is equivalent to maximizing the disruption effect of the protective perturbation under the transformations:
\begin{equation}
\label{equ: policy optimization}
\arg\max_{\theta}
\mathbb{E}_{t \in S_v}
\Big\|
\mathcal{F}\!\big(t(x+\delta)\big)
-
\mathcal{F}(x)
\Big\|_2^2.
\end{equation}
Note that $S_v$ is distinct from the test transformation set, $S_t$, since the benign users have no access to the latter. Ideally, $S_v$ should be selected to approximate $S_t$ well so that the transformation policy optimized against the $S_v$ can generalize well to $S_t$. 

Unfortunately, the entire pipeline from $\theta$ to the objective \cref{equ: policy optimization} is not fully differentiable so that the optimization cannot be easily solved through backpropagation. We adopt REINFORCE algorithm \cite{williams_simple_1992} to update our policy model as explained below.

Given a clean source image $x$, we sample $T$ trajectories of transformations and generate $T$ perturbed images $\{x_{adv}^{(j)}\}_{j=1}^{T}$ using \cref{equ:pgd_xadv_short} with $N=1$ for efficiency. Each trajectory draws $L$ transformations from the current transformation policy, $\pi_{S_p}(x; \theta)$. To prevent the policy from collapsing onto a small subset of transformations, a probability cap is applied during sampling, ensuring that no single transformation receives excessively large probability mass.

For each trajectory $j$, all validation transformations 
$t(\cdot)\in\mathcal{S}_{\text{val}}$ 
are applied to $x_{adv}^{(j)}$,
and the mean validation loss is computed as
\begin{equation}
r_j =
\frac{1}{|\mathcal{S}_{v}|}
\sum_{t\in\mathcal{S}_{v}}
\Big\|
\mathcal{F}\!\big(t(x+\delta_j)\big)
-
\mathcal{F}(x)
\Big\|_2^2.
\end{equation}

These validation losses $\{r_j\}_{j=1}^{T}$ serve as trajectory-level rewards for the policy network.
To stabilize training and enable both positive and negative updates, we apply baseline trick as
\begin{equation}
\bar{r} = \frac{1}{T}\sum_{j=1}^{T} r_j, 
\qquad
\hat{r}_j = r_j - \bar{r}.
\end{equation}

The policy parameters are updated 
by minimizing the negative correlation between the rewards 
and the log-probabilities of sampled transformations:
\begin{equation}
\mathcal{L}_{\text{policy}}
=
-\frac{1}{T}
\sum_{j=1}^{T}
\hat{r}_j \log p_j,
\end{equation}
where $p_j$ denotes the probability of the sampled trajectory, \ie the product of probabilities of sampled $L$ transformations.
Minimizing $\mathcal{L}_{\text{policy}}$ increases the likelihood of trajectories with positive advantages ($\hat{r}_j>0$) and suppresses those with negative ones.
The parameters are then optimized using stochastic gradient descent (SGD):
\begin{equation}
\theta \leftarrow 
\theta - \eta\, \nabla_\theta \mathcal{L}_{\text{policy}},
\end{equation}
where $\eta$ is the learning rate.

\section{Experiment}


\begin{table*}[tbp]
\centering
\caption{\textbf{The performance of our methods under no and various transformations on the FFHQ dataset}. ``Train'' and ``Test'' refer to the training and test splits of the image data, representing seen (by the policy model) and unseen data scenarios, respectively. ``Overall'' averages the performance over all test transformations. Performance is evaluated using the ID similarity metric, with the best results in each setting \textbf{highlighted}.}
\label{tab:table1}
\resizebox{\textwidth}{!}{%
\begin{tabular}{@{}llcccccccc@{}}
\toprule
\multirow{2}{*}{Data}       & \multirow{2}{*}{Method} & \multirow{2}{*}{\makecell[c]{No \\ Transformation}} & \multicolumn{7}{c}{Transformations}                                                                                  \\ \cmidrule(l){4-10} 
                        &                         & \multicolumn{1}{c}{}                                                                              & Noise          & Color-space    & Blur           & Stylization    & Compression    & Geometric      & Overall        \\ \midrule
\multirow{5}{*}{Train}   & No attack               & 0.540                                                                                                   & 0.503          & 0.513          & 0.551          & 0.485          & 0.498          & 0.441          & 0.496          \\
                        & PGD                     & 0.191                                                                                                  & 0.359          & 0.312          & 0.526          & 0.306          & 0.373          & 0.366          & 0.364          \\
                        & DF-RAP                 &  0.273                                                                                                 & 0.386          & 0.345          & 0.490          & 0.322          & 0.354          & 0.371          & 0.373          \\
                        & PGD-EOT           &  0.308                                                                                                 & 0.365          & 0.348          & 0.304          & 0.317          & 0.327          & 0.299          & 0.328          \\ \cmidrule(l){2-10} 
                        & Ours                    & \textbf{0.180}                                                                                                  & \textbf {0.260}    & \textbf{0.243}    & \textbf{0.242}    & \textbf{0.214}    & \textbf{0.241}    & \textbf {0.261}    & \textbf{0.242}    \\ \midrule
\multirow{5}{*}{Test} & No attack               & 0.534                                                                                                  & 0.495          & 0.494          & 0.529          & 0.473          & 0.498          & 0.438          & 0.482          \\
                        & PGD                     &  0.192                                                                                                 & 0.355          & 0.310          & 0.521          & 0.304          & 0.366          & 0.363          & 0.360          \\
                        & DF-RAP                 &  0.284                                                                                                 & 0.387          & 0.350          & 0.489          & 0.326          & 0.359          & 0.371          & 0.375          \\
                        & PGD-EOT           &  0.304                                                                                                 & 0.359          & 0.342          & 0.297          & 0.312          & 0.322          & 0.293          & 0.322          \\ \cmidrule(l){2-10} 
                        & Ours                    & \textbf{0.181}                                                                                                  & \textbf{0.258} & \textbf{0.239} & \textbf{0.234} & \textbf{0.212} & \textbf{0.236} & \textbf{0.255} & \textbf{0.238} \\ \bottomrule
\end{tabular}%
}
\end{table*}

\subsection{Experimental Setup}

\textbf{Choices of Transformations.} To evaluate the robustness of adversarial examples, we adopt a collection of image transformations inspired by Sitawarin et al.\cite{sitawarin_demystifying_2022}. The objective is to identify transformations that can effectively destroy adversarial perturbations while introducing only minimal degradation to clean images. Such transformations pose a significant threat in practical settings, as they may arise naturally from environmental factors or be deliberately induced by an adversary to neutralize adversarial protections. To better capture the diversity of real-world perturbations, several transformations are slightly modified or extended. In total, we curate a set of $K = 30$ commonly used transformations, grouped into 6 categories, with the detailed list provided in Sec.~\ref{sec:sensitivity_simswap}.

\textbf{Implementation Details.} Following prior face-swapping research \cite{baliah2025realistic}, 1,000 source images and 1,000 target images from FFHQ \cite{karras2019style} are used to construct the training facial image set. For testing, we use another 1,000 source–target pairs from FFHQ. All images are resized to $256 \times 256$ before processing. Identity features are extracted using ArcFace \cite{deng2019arcface}, and identity consistency (\textbf{ID similarity}) is assessed by computing the cosine similarity between the ID embeddings of the face-swapped images and those of the corresponding source images. The PGD parameters used in the perturbation generation stage are 
a number of $N = 150$ iterations, a step size of $\alpha = 0.01$, 
and an $\ell_\infty$ perturbation budget of $\epsilon = 0.05$. For consistency, the FGSM also adopts the same perturbation budget 
$\epsilon = 0.05$.
All the experiments were conducted on an HPC server running a Linux environment equipped with an Intel(R) Xeon(R) Gold 6330 CPU, and an NVIDIA A100 GPU with 40\,GB of memory.
The full set of hyperparameters is provided in Appendix B.

\textbf{Baseline methods.} The proposed PGD-EOLT is compared to four baselines, No attack, PGD \cite{madry2017towards}, DF-RAP \cite{qu_df-rap_2024}, and PGD-EOT \cite{athalye_synthesizing_2018}.

\subsection{Improved Robustness against Transformations}

We first evaluate our method in a setup where all transformations are available for both perturbation generation and evaluation. This simulates a scenario where users and attackers have access to the same set of transformations. For the seen data setup, we use the training split of FFHQ for both policy learning and evaluation, meaning the source and target images used for face swapping are seen by the transformation policy model during training. For the unseen data setup, we use the training split for policy learning and the test split for evaluation, allowing us to assess the generalization ability of the learned policy model.






As shown in Tab.~\ref{tab:table1}, our method achieved the highest robustness against 6 transformation categories, under both seen and unseen settings in the FFHQ dataset. In general, it achieved an average ID similarity score of 0.242 in the seen setting, surpassing the second best method, PGD-EOT, with a relative increase of 26.2\% . In the unseen setting, our method achieved an average ID similarity of 0.238, a relative increase of 26.1\% in robustness compared to the second best method. Our method demonstrated the highest performance gain under the Stylization transformation, a relative increase of 30.1\% and 30.3\% compared to the second best method in the seen and unseen settings, respectively. For geometric transformation, the improvement in robustness is less prominent compared to other transformation categories, but our method still achieved a relative increase of 12.7\% and 13.0\% in the seen and unseen settings compared to the second best method. Notably, the performance gap of our method between seen and unseen data is marginal, indicating that our learned policy network can well generalize to unseen data.

It is also noteworthy that our method outperforms the PGD baseline even under the no-transformation setting, despite not being explicitly optimized for this scenario. Among all the robustness-focused methods compared, ours is the only one to achieve this. This indicates that our approach strikes a better balance between clean and robust performance.

\subsection{Generalization to Unseen Transformations}
\label{generalization to unseen trans}
This section evaluates the generalization ability of our method to unseen transformations, where the transformation set used for perturbation generation has no overlap with the set used for evaluation. This setup simulates a scenario in which benign users have no knowledge of the specific transformations that may be applied by an attacker.
Two specific setups are tested: intra-category and inter-category. In the intra-category setup, one transformation per category is used for validation, and the remaining transformations are evenly divided into training and test sets (12 each). In the inter-category setup, the 6 categories are grouped into three disjoint pairs, each assigned to a different split—training, validation, or test—ensuring no category overlap between splits.

\begin{table}[t]
\centering
\caption{\textbf{The performance of our methods under seen and unseen transformations on the train split of FFHQ dataset.} In the intra-category setting, transformations are split into Train/Val/Test within each category, while in the inter-category setting, transformations are split across categories, with no overlap between training, validation, and test sets.}
\label{tab: unseen transformations}
\resizebox{\columnwidth}{!}{%
\begin{tabular}{@{}llccccc@{}}
\toprule
\multirow{2}{*}{Setup}              & \multirow{2}{*}{Method}  & \multirow{2}{*}{\makecell[c]{No \\ Trans.}} & \multicolumn{4}{c}{Transformations} \\ 
    \cmidrule(l){4-7} 
                        &                         & \multicolumn{1}{c}{}                                                                             & Train          & Val.     & Test           & Overall        \\ \midrule

\multirow{5}{*}{\makecell[c]{Intra- \\ category}} & No attack & 0.540    & 0.509          & 0.420          & 0.518          & 0.496          \\
                   & PGD  & 0.191         & 0.337          & 0.330          & 0.393          & 0.364          \\
                   & DF-RAP & 0.273      & 0.352          & 0.314          & 0.408          & 0.373          \\
                   & PGD-EOT & 0.353 & 0.348          & 0.297          & 0.380          & 0.355          \\ \cmidrule(l){2-7} 
                   & Ours  & \textbf{0.187}        & \textbf{0.228} & \textbf{0.212} & \textbf{0.281} & \textbf{0.252} \\ \midrule                   
\multirow{5}{*}{\makecell[c]{Inter- \\ category}} & No attack  & 0.540   & 0.509          & 0.475          & 0.507          & 0.496          \\
                   & PGD  & \textbf{0.191}         & 0.386          & 0.363          & 0.335          & 0.364          \\
                   & DF-RAP & 0.273      & 0.383          & 0.379          & 0.349          & 0.373          \\
                   & PGD-EOT &0.342 & 0.328          & 0.356          & 0.364          & 0.348          \\ \cmidrule(l){2-7} 
                   & Ours  & 0.226        & \textbf{0.233} & \textbf{0.299} & \textbf{0.280} & \textbf{0.270} \\ 
                    \bottomrule
\end{tabular}%
}
\vspace{-4mm}
\end{table}




Tab.~\ref{tab: unseen transformations} summarizes the results. Our method achieved the best performance with and without image transformations in the intra-category setting, and compared to the baselines, it significantly improved the robustness under all splits. Its overall ID similarity score is 29.0\% higher than the second best method when images are transformed. In the inter-category setting, our method also achieved the best performance when images are transformed, with a relative overall increase of 22.4\% in robustness compared to the second best method. When there is no transformation, in the inter-category setting, our method still demonstrated robustness, and ranked as the second best among all methods. 
Overall, these empirical results suggest that our method generalize well to unseen transformations.




\subsection{Ablation Study}
\label{ablation}

We further conducted ablation studies to investigate the effectiveness of different configurations of our method in improving robustness. 

\begin{figure}[t]
    \centering
    \includegraphics[width=.8\linewidth]{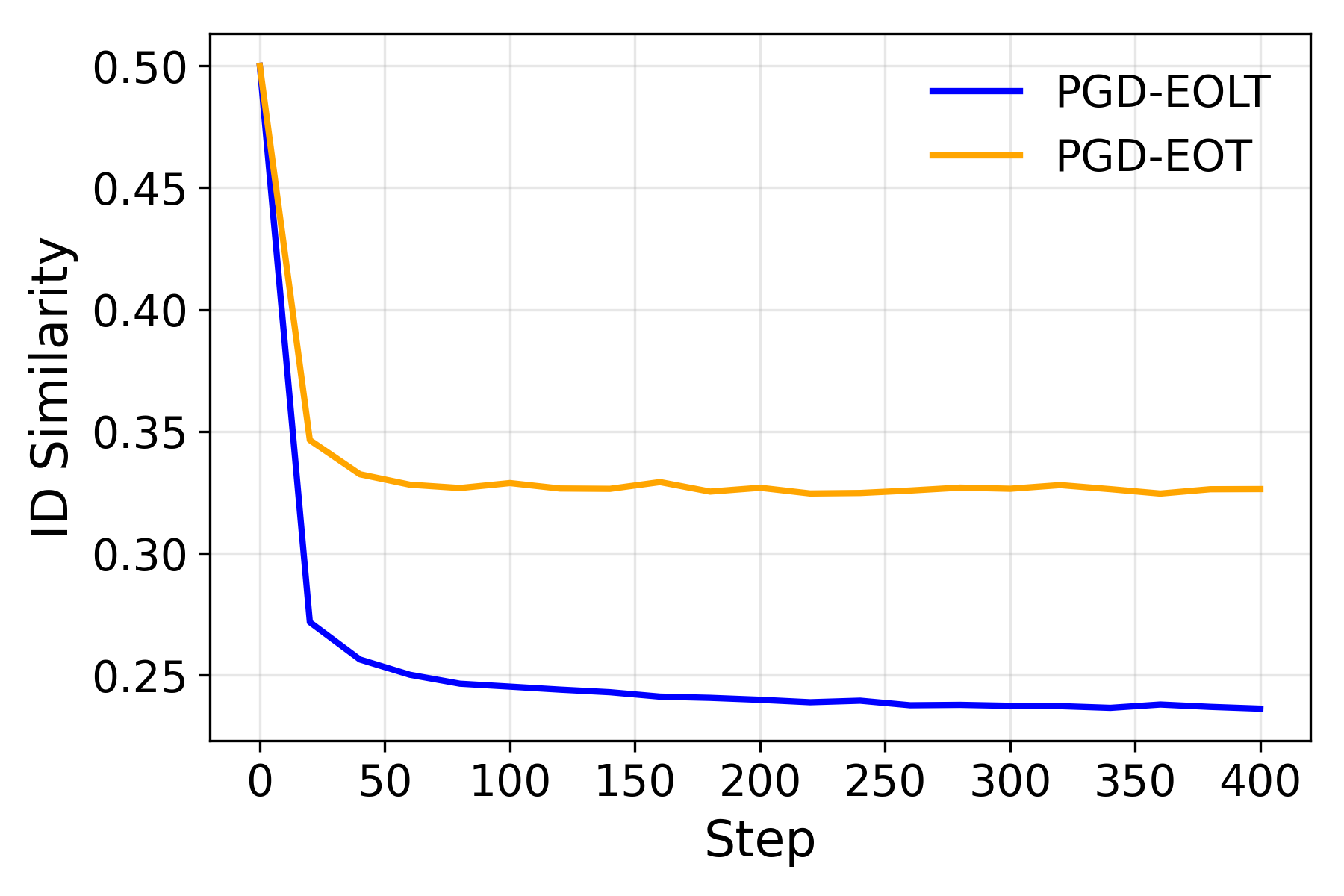}
    \caption{\textbf{The performrance of PGD-EOT and our method PGD-EOLT with the number of steps used in PGD on the train split of FFHQ dataset.} The figure highlights distinct optimization dynamics, with PGD-EOT stabilizing quickly while PGD-EOLT continues improving over extended iterations.}
    \label{fig:pgdsteps}
\end{figure}

\textbf{Convergence of PGD Attack.}
As shown in Fig.~\ref{fig:pgdsteps}, PGD-EOT converges after 50 steps, whereas ours PGD-EOLT benefits with more steps of optimization, and eventually converges after around 150 steps. Hence, our selection of PGD with 150 steps ensures that both methods converge for a fair comparison on adversarial effectiveness.

\begin{figure*}[tbp]
    \centering
    \includegraphics[width=\linewidth]{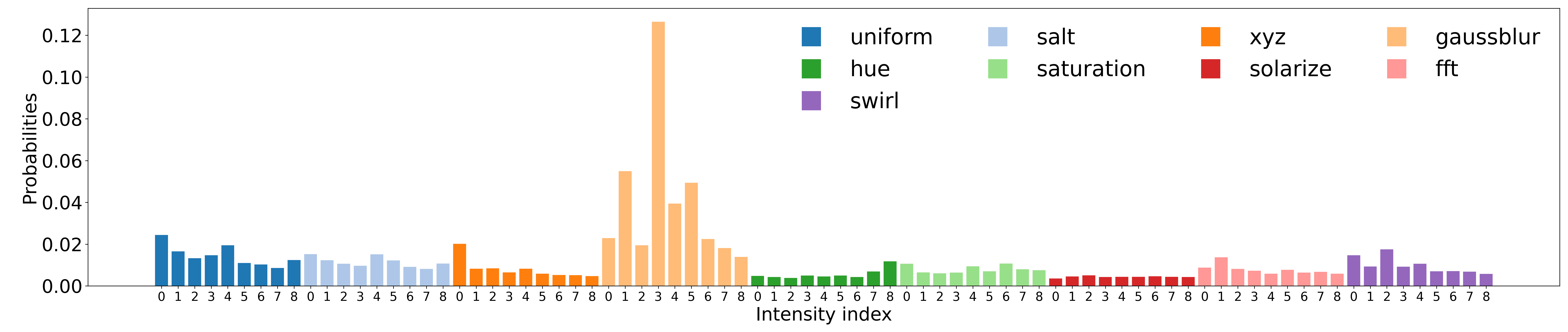}
    \caption{\textbf{Learned probability distribution over sub-policies available for perturbation generation averaged over all images in the train split of FFHQ dataset.} There are 81 sub-policies in total. Each color corresponds to 1 of the 9 training transformations in this setup, and each transformation is associated with 9 discrete intensity levels, indexed from 0 to 8 in the figure.} 
    \label{fig: distribution}
    \vspace{-4mm}
\end{figure*}

\textbf{Policy model architecture.} Tab.~\ref{tab:arch_table} presents the results of our method with different backbone architectures, Vision Transformer (ViT-B/16) \cite{dosovitskiy2020image}, WideResNet28-10 (WRN28-10) \cite{zagoruyko2016wide}, and PreAct ResNet-18 (PRN18) \cite{he2016identity}. The results show that the choice of policy model architecture has no considerable impact on the final performance of our method in both with and without transformation settings.

\begin{table}[t]
\centering
\caption{\textbf{The performance of our methods with different network architectures as the policy model backbone on the train split of FFHQ dataset.} Color / Styl. / Comp. / Geo. denotes Color-space / Stylization / Compression / Geometric, respectively.}
\label{tab:arch_table}
\resizebox{\columnwidth}{!}{
\begin{tabular}{lcccccccc}
\toprule
\multirow{2}{*}{Architecture}   & \multirow{2}{*}{\makecell[c]{No \\ Trans.}} & \multicolumn{7}{c}{Transformations} \\ 
    \cmidrule(l){3-9} 
                        &                          \multicolumn{1}{c}{}                                                                             & Noise & Color & Blur & Styl. & Comp. & Geo. & Overall        \\ \midrule

ViT-B/16     & 0.181 & 0.261 & 0.240 & 0.246 & 0.214 & 0.239 & 0.253 & 0.241 \\
WRN28-10     & \textbf{0.179} & 0.275 & 0.242 & 0.242 & 0.214 & 0.242 & \textbf{0.250} & 0.243 \\
ResNet18  & 0.181 & \textbf{0.258} & \textbf{0.239} & \textbf{0.234} & \textbf{0.212} & \textbf{0.236} & 0.255 & \textbf{0.238} \\
\bottomrule
\end{tabular}
}
\vspace{-3mm}
\end{table}

\begin{table}[tbp]
\centering
\caption{\textbf{The performance of our methods with different initial learning rates for training the policy model on the train split of FFHQ dataset.} Color / Styl. / Comp. / Geo. denotes Color-space / Stylization / Compression / Geometric, respectively.}
\label{tab:lr_table}
\resizebox{\columnwidth}{!}{
\begin{tabular}{lcccccccc}
\toprule
\multirow{2}{*}{Init. LR}   & \multirow{2}{*}{\makecell[c]{No \\ Trans.}} & \multicolumn{7}{c}{Transformations} \\ 
    \cmidrule(l){3-9} 
                        &                          \multicolumn{1}{c}{}                                                                             & Noise & Color & Blur & Styl. & Comp. & Geo. & Overall        \\ \midrule
0.1    & 0.203 & 0.280 & 0.260 & 0.265 & 0.232 & 0.269 & 0.282 & 0.263 \\
0.01   & 0.183 & 0.260 & 0.245 & 0.237 & 0.216 & 0.241 & 0.264 & 0.243 \\
0.001  & \textbf{0.181} & \textbf{0.258} & \textbf{0.239} & \textbf{0.234} & \textbf{0.212} & \textbf{0.236} & 0.255 & \textbf{0.238} \\
0.0001 & \textbf{0.181} & 0.280 & 0.244 & 0.252 & 0.216 & 0.246 & \textbf{0.252} & 0.247 \\
\bottomrule
\end{tabular}
}
\end{table}

\begin{figure}[tbp]
    \centering
    \includegraphics[width=.9\linewidth]{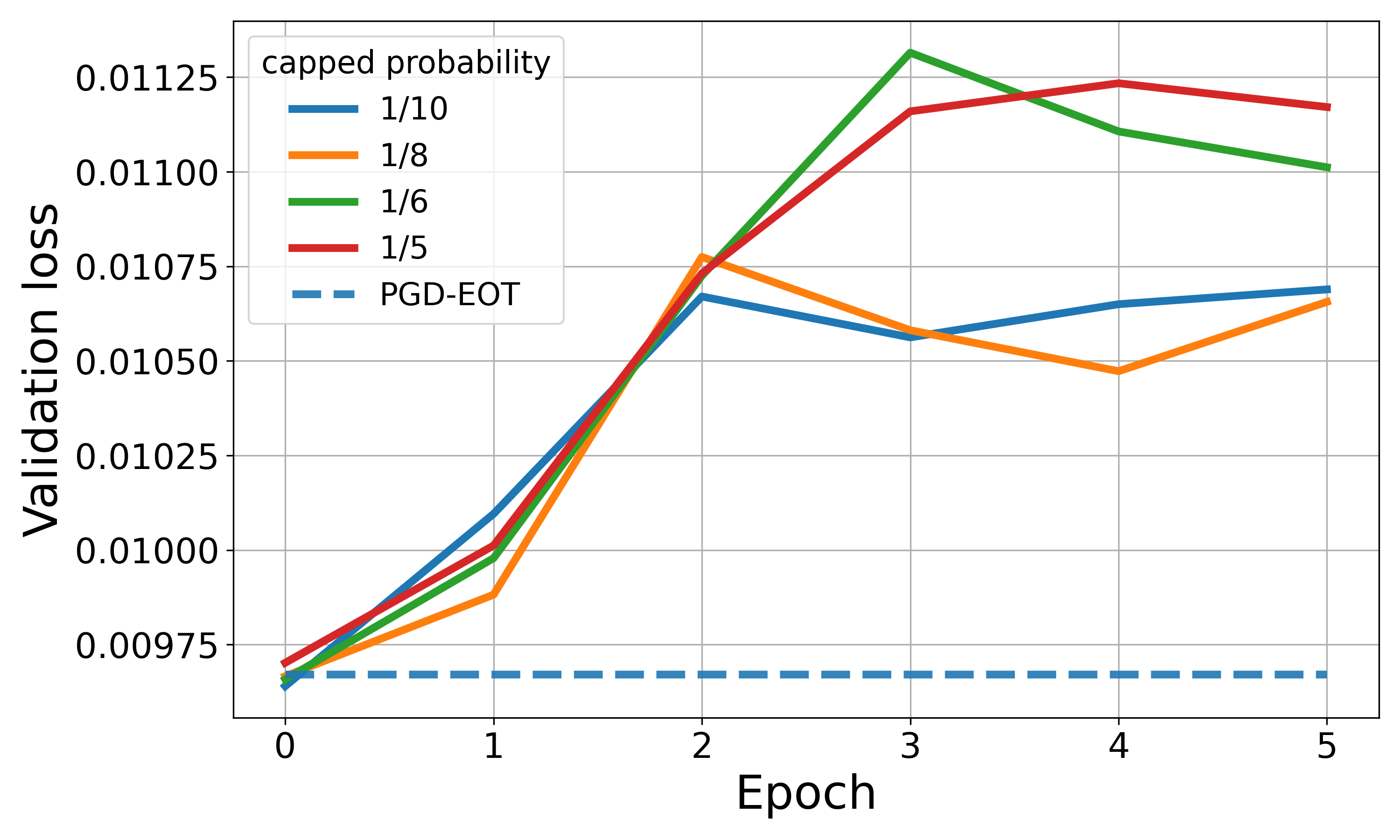}
    \caption{\textbf{The change of validation loss of our method with different capped sampling probabilities over the course of training.} Each solid line corresponds to a manually specified capped probability value (1/10, 1/8, 1/6, 1/5). This value serves as an upper bound that the sampling probability of any sub-policy is not allowed to exceed during training. Higher validation loss corresponds to stronger attack success, so curves with higher loss indicate more effective protective perturbation. The dashed line shows validation loss of the PGD-EOT baseline. 
    }
    \label{fig: losscurve}
\end{figure}

\textbf{Policy learning rate.} We further studied the impacts of different initial learning rates on the final robustness. As shown in Tab.~\ref{tab:lr_table}, a moderately small initial learning rate helps achieve better performance, but the final performance does not benefit from further smaller learning rate.


\subsection{Learning Analysis}



This section analyzes the results of policy learning to show that our policy model successfully learns a transformation policy that differs from the uniform distribution and is more effective in generating stronger protective perturbations.

\textbf{Learned distribution.}
Fig.~\ref{fig: distribution} shows the probability distribution learned by the policy network. We observe that the policy assigns substantially higher probabilities to \texttt{gaussblur} and \texttt{uniform}, while \texttt{solarize} and \texttt{hue} receive near-zero weights, which directly aligns with Fig.~\ref{fig:simswap_heatmap}, where blur and noise families exhibit strong cross-transformation generalization, whereas \texttt{hue} and stylization transformations show poor transferability or even degrade robustness. This learned allocation pattern validates our analytical findings and demonstrates that the policy network successfully learns to prioritize beneficial transformations while automatically downweighting toxic ones, confirming the effectiveness of adaptive sampling over uniform random sampling.


\textbf{Loss curve.} As shown in Fig.~\ref{fig: losscurve}, we evaluate four capped-probability settings under the same training configuration. Across training, the PGD-EOLT validation loss curves consistently lie above the uniform PGD-EOT baseline and display a steadily increasing trend. This indicates that the learned non-uniform policy progressively improves over training, steering our method toward perturbations that better withstand image transformations and therefore produce stronger protection.

Last, we also visualize some examples of our protection results in Appendix C.

\section{Conclusion}
In this work, we addressed the critical challenge of generating transformation-robust protective perturbations against deepfake face swapping. Through comprehensive analysis of 30 transformations, we revealed that transformations exhibit vastly different generalization behaviors and identified defensive bottlenecks that fundamentally limit uniform sampling approaches. To overcome these limitations, we proposed EOLT, the first framework to treat transformation distribution as a learnable component, enabling automatic prioritization of critical transformations and adaptive perturbation generation tailored to each input image. Extensive experiments demonstrate that our method achieves substantial robustness improvements, outperforming state-of-the-art methods by 26\% on average while maintaining strong generalization to unseen transformations and data.

\clearpage
{
    \small
    \bibliographystyle{ieeenat_fullname}
    \bibliography{ref, references}
}

\clearpage
\setcounter{page}{1}
\maketitlesupplementary

\appendix
\section*{Overview}
This supplementary material provides additional results and analyses that further support the findings presented in the main paper. The contents are organized as follows:
\begin{itemize}
    \item Section~\ref{A} presents the full cross-transformation identity similarity matrix.
    \item Section~\ref{B} summarizes the hyperparameters used in each experimental setup.
    \item Section~\ref{C} includes visualization results and additional learned distributions of our method.
\end{itemize}

\section{Cross-Transformation Generalization}
\label{A}
\begin{figure*}
    \centering
    \includegraphics[width=1.0\linewidth]{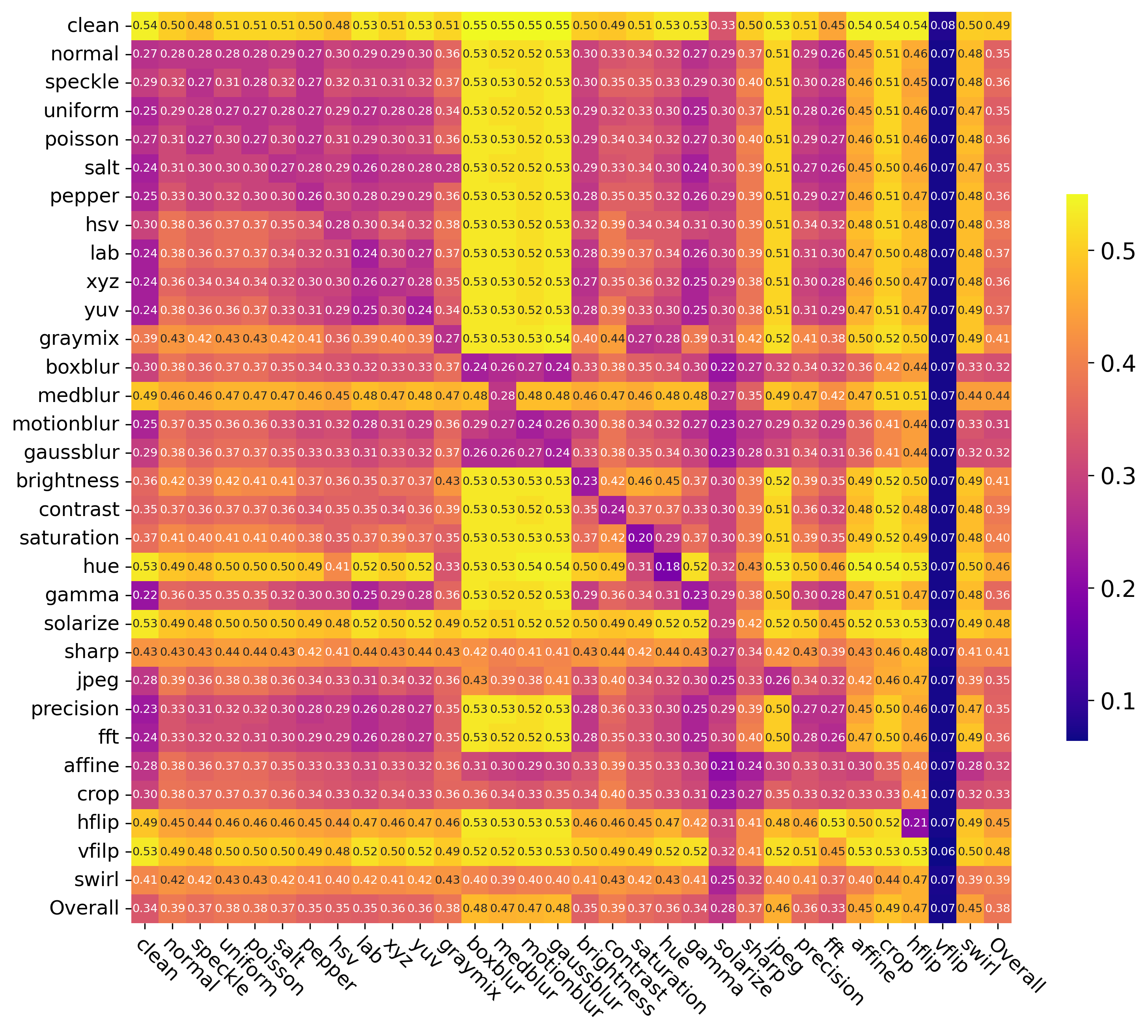}
    \caption{\textbf{Cross-transformation identity similarity matrix over all 32 transformations.} Rows indicate transformations used during perturbation generation, while columns indicate transformations applied at test time. ``Clean'' denotes no transformation applied.}
    \label{fig:fullheatmap}
\end{figure*}

The full 32 $\times$ 32 cross-transformation identity similarity matrix in the Fig.~\ref{fig:fullheatmap} provides a more global view of how different transformations interact.

\subsection{Behaviors of Transformations used in Training}
\label{Behaviors of Transformations when used in Training}
Noise-type transformations (\texttt{normal}, \texttt{uniform}, \texttt{speckle}, \texttt{poisson}, \texttt{salt}, \texttt{pepper}) and blur-type transformations (\texttt{boxblur}, \texttt{motionblur}, \texttt{gaussblur}; excluding \texttt{medblur}) exhibit strong within-family generalization (all within 0.32), and these three blur transformations further generalize well to most other transformation categories. In contrast, perturbations trained on \texttt{medblur}, \texttt{hue}, \texttt{solarize}, \texttt{sharp}, \texttt{hflip}, \texttt{vflip}, and \texttt{swirl} exhibit ID similarities above 0.45 for most off-diagonal entries, failing to transfer to unseen transformations.

\subsection{Behaviors of Transformations when Tested on}
\label{Behaviors of Transformations when Tested on}
\texttt{vflip} yields extremely low identity similarity (0.07) due to its drastic geometric distortion of the facial image. Because of this outlier behavior, it can be disregarded when interpreting broader trends. \texttt{solarize} exhibits a similar identity-disrupting effect for the same reason, though its impact is milder.

Blur (\texttt{boxblur},\ \texttt{motionblur},\ \texttt{gaussblur},\ \texttt{medblur},) and Geometric (\texttt{affine},\ \texttt{crop},\ \texttt{hflip},\ \texttt{swirl}) families remain high ID similarities (over 0.52 for blur and over 0.45 for geometric in most cases) when the training perturbations come from outside their respective families, reflecting their resistances to being generalized from other transformations.

In contrast, noise, color-space, and stylization transformations are better covered by perturbations trained on other transformations, except for the poorly generalizing cases identified in Appendix~\ref{Behaviors of Transformations when used in Training}. Within the compression family, \texttt{jpeg} is much harder to generalize to (most values are above 0.5), while \texttt{fft} and \texttt{precision} are clearly better.

\section{Hyperparameters}
\label{B}

\begin{figure*}
    \centering
    \includegraphics[width=0.81\linewidth]{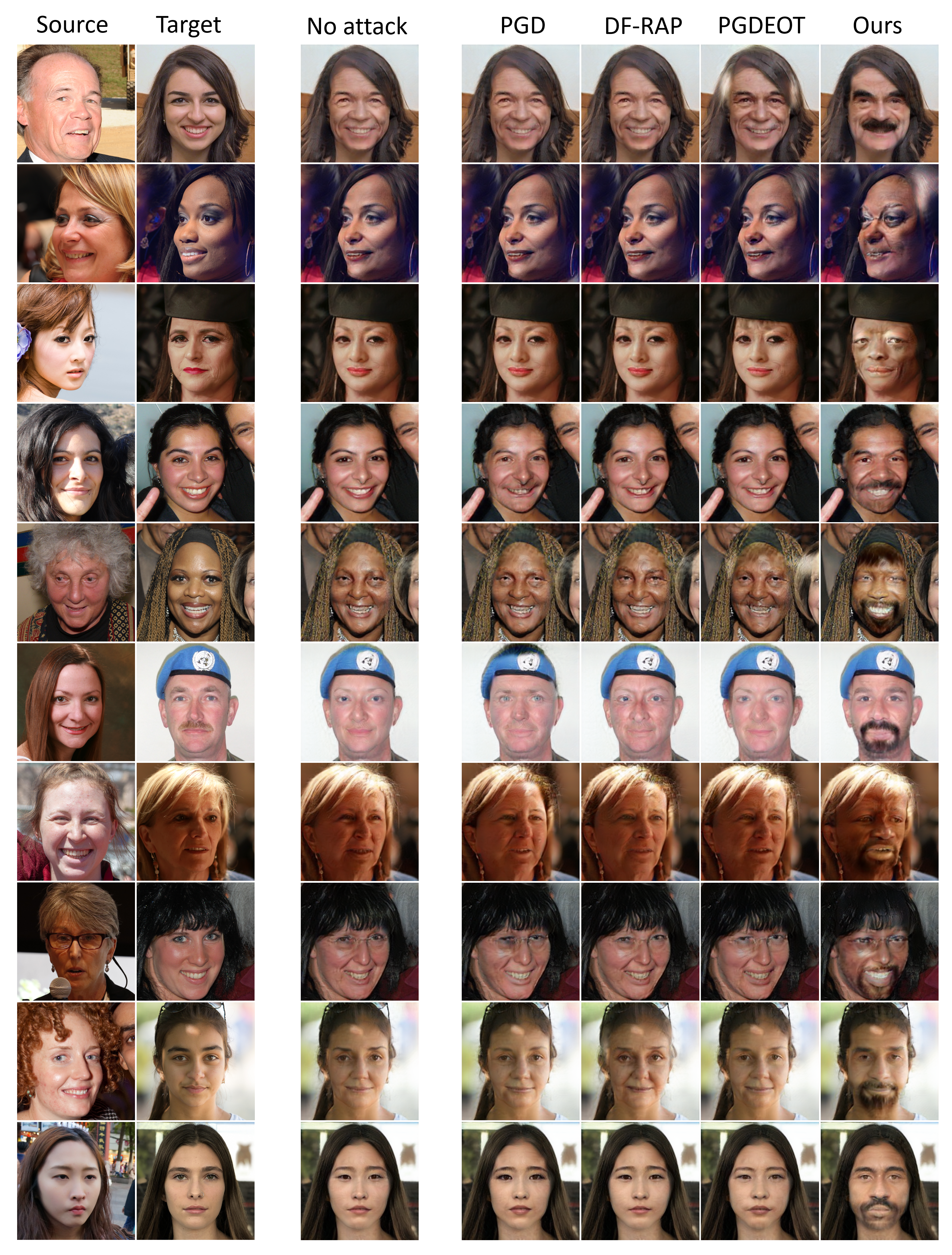}
    \caption{\textbf{The visualization of our protection results.} Each row shows a source--target pair (first two columns) followed by fake images produced under no attack and several attack methods, with a single transformation applied (including no attack and all attack methods) before face swapping. The transformation used for ten pairs from top to bottom are: \texttt{boxblur}, \texttt{boxblur}, \texttt{boxblur}, \texttt{brightness}, \texttt{crop}, \texttt{fft}, \texttt{gamma}, \texttt{hflip}, \texttt{normal}, \texttt{speckle}.}
    \label{fig:visualization}
\end{figure*}

\begin{table}
\centering
\caption{Shared hyperparameters}
\label{tab:hyperparams}
\resizebox{\columnwidth}{!}{%
\begin{tabular}{l c}
\toprule
\textbf{Parameter} & \textbf{Value} \\
\midrule
Number of epochs          & 5 \\
Batch size                & 8 \\
Optimizer                 & SGD (momentum = 0.9) \\
Learning rate             & 0.001 \\
Scheduler                 & Warmup + cosine decay \\
Warmup epochs             & 2 \\
Trajectory length $K$     & 6 \\
Trajectories per sample $L$  & 10 \\
PGD step size $\alpha$            & 0.01 \\
PGD/FGSM max perturbation $\epsilon$ & 0.05 \\
Capped probability        & $1/6$ \\
Policy architecture        & PreAct ResNet-18 \\
\bottomrule
\end{tabular}%
}
\end{table}




\begin{figure*}
    \centering
    \includegraphics[width=\linewidth]{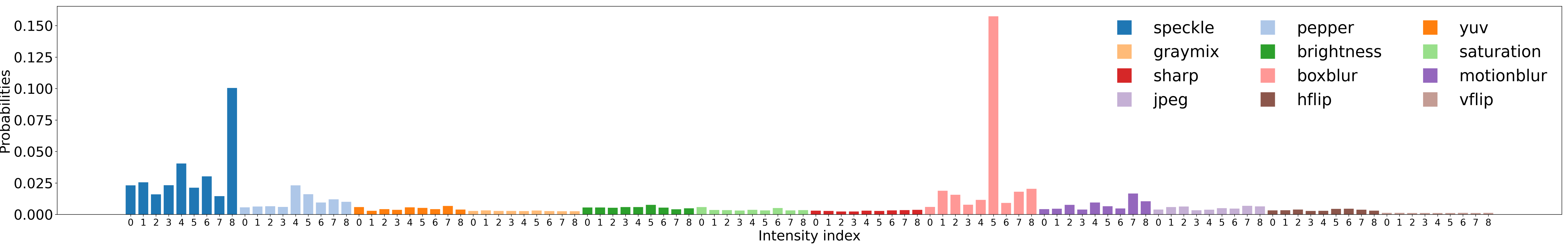}
    \caption{\textbf{Learned probability distribution over the 108 sub-policies under the intra-category setup.} The distribution is computed over all sub-policies available for perturbation generation and averaged across all images in the FFHQ training split. Each color corresponds to 1 of the 12 training transformations in this setup, and each transformation is associated with 9 discrete intensity levels, indexed from 0 to 8 in the figure.}
    \label{fig:926}
    \vspace{4mm}

    \includegraphics[width=\linewidth]{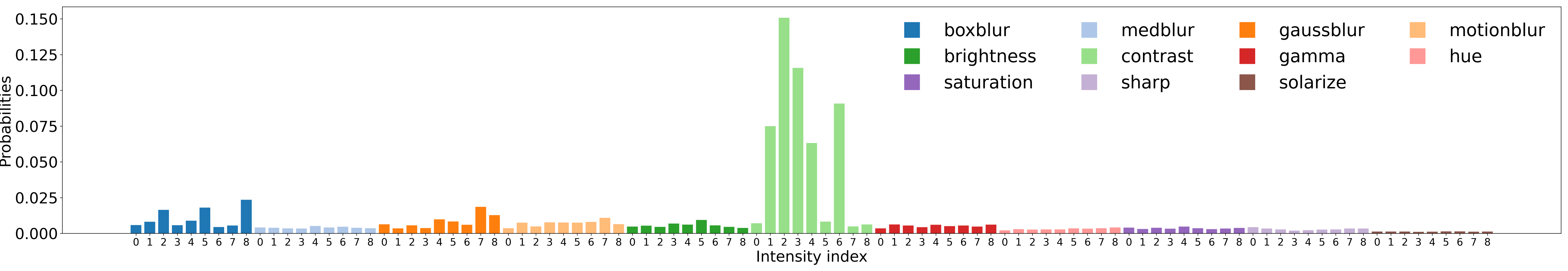}
    \caption{\textbf{Learned probability distribution over the 99 sub-policies under the inter-category setup.} The meaning of intensity indices and the averaging procedure is identical to Fig.~\ref{fig:926}, except that this setup contains 11 training transformations.}
    \label{fig:1017}
\end{figure*}

\subsection{Shared parameters}

The PGD-related hyperparameters are taken from the DF-RAP method \cite{qu_df-rap_2024}. In contrast, the hyperparameters used for policy training are tuned specifically for our setting. Increasing the batch size or the trajectory length $K$ or the number of trajectories per sample often leads to more stable optimization and can improve training performance. However these choices also bring significantly higher computational cost. With the configuration reported in Tab.~\ref{tab:hyperparams} the policy model already provides a substantial improvement over the baseline and converges stably within $5$ epochs as shown in Fig.~\ref{fig: losscurve}. The capped probability also plays an important role. A smaller value reduces the differences among sampled sub-policies and a larger value causes the sampling distribution to collapse. Both situations reduce robustness. Experiments show that a capped probability of $1/6$ provides a stable balance and this behavior is illustrated in Fig.~\ref{fig: losscurve}. Further details regarding the learning rate schedule and the policy architecture are provided in Sec.~\ref{ablation}.

\subsection{All-seen}

The all-seen setting uses the complete set of $30$ transformations for training validation and testing. This configuration is adopted in the main experiments to assess the full capacity of the learned transformation policy. Since every transformation is observable during training the policy can adapt to the entire transformation space without encountering unseen cases at evaluation time.

\subsection{Intra-category}
\label{intra}
The intra-category setting evaluates the ability of the policy to handle unseen transformations that belong to the same semantic families. In our case, one transformation from each of the six categories is assigned to the validation split (\texttt{poisson}, \texttt{lab}, \texttt{medblur}, \texttt{hue}, \texttt{fft}, \texttt{affine}) and the remaining transformations in each category are evenly divided into the training (\texttt{speckle},\ \texttt{pepper},\ \texttt{yuv},\ \texttt{graymix},\ \texttt{boxblur},\ \texttt{motionblur},\ \texttt{brightness},\ \texttt{saturation},\ \texttt{sharp},\ \texttt{jpeg},\ \texttt{hflip},\ \texttt{vflip}) and test (\texttt{normal},\ \texttt{uniform},\ \texttt{salt},\ \texttt{hsv},\ \texttt{xyz},\ \texttt{gaussblur},\ \texttt{contrast},\ \texttt{gamma},\ \texttt{solarize},\ \texttt{precision},\ \texttt{crop},\ \texttt{swirl}) splits. This results in $12$ training transformations, $6$ validation transformations, and $12$ test transformations. This setup tests whether the policy can generalize to novel transformations that share structural similarity with those observed during training.

\subsection{Inter-category}
\label{inter}
The inter-category setting examines whether the policy can generalize across entirely different transformation families. In our setup, the six categories are grouped into three disjoint pairs: the training split uses Blur (4) and Stylization (7), the validation split uses Noise (6) and Geometric (6), and the test split uses Color-space (5) and Compression (3). Here, the numbers in parentheses indicate the number of transformations contained in each category. This configuration ensures that none of the categories included during training appear in either the validation or test splits. Consequently, it evaluates whether the policy can transfer to transformation families that are semantically distinct from those encountered during training, representing a significantly more challenging unseen-transformation scenario.

\section{Visualization}
\label{C}
\subsection{Attack Outcome Examples}
We provide ten pairs of qualitative results in Fig.~\ref{fig:visualization}. For each pair, the first two columns show the source and target images. The third column presents the fake images produced without any attack, serving as a clean reference. The remaining four columns correspond to different attack methods, where each attack perturbs the source image and the perturbed result is then passed through a \emph{single transformation} before face swapping with the target image. 

As shown, our method consistently demonstrates strong robustness under all transformations, producing fake images that clearly reveal the degradation introduced by the perturbations. In contrast, the fake images produced by the other attack methods remain visually almost indistinguishable from the \emph{no attack} results, revealing their weak resilience to transformation-induced distortions.

\subsection{Learned Distributions}
We further present the learned distributions for the two groups of experiments. The experimental setups correspond to Fig.~\ref{fig:926} for Appendix~\ref{intra} and Fig.~\ref{fig:1017} for Appendix~\ref{inter}. Across extensive trials, we observe a clear pattern in the intra-category setting: when both the training and validation sets contain blur-type transformations (except for the case where the training set includes only \texttt{medblur}), the learned distribution consistently increases the sampling probability of blur transformations. This aligns with the conclusion in Appendix~\ref{Behaviors of Transformations when Tested on}, namely that achieving good generalization to blur-type transformations in the validation set requires sufficient blur coverage in the training set, which in turn also improves the overall generalization ability of the training distribution.

Notably, the learned distributions differ significantly from the uniform distribution assumed in standard EOT, indicating that the policy model captures meaningful, non-uniform preferences, rather than treating all transformations equally. This further demonstrates that our optimization procedure effectively identifies transformation–intensity combinations that contribute more to robustness.

\end{document}